\documentclass[11pt]{article}
\usepackage{rotating}
\usepackage{acl2015}
\usepackage{times}
\usepackage{url}
\usepackage{latexsym}
\usepackage{tabularx}
\usepackage{makecell}
\usepackage{hyperref}
\usepackage{multirow}
\usepackage{multicol}
\usepackage{booktabs}
\usepackage{amsmath}
\DeclareMathOperator*{\argmax}{argmax}
\usepackage{appendix}
\usepackage{CJKutf8}
\usepackage[section]{placeins}
\usepackage[hang,flushmargin]{footmisc} 
\usepackage[T1]{fontenc}
\usepackage{lipsum}
\newcommand\blfootnote[1]{%
  \begingroup
  \renewcommand\thefootnote{}\footnote{#1}%
  \addtocounter{footnote}{-1}%
  \endgroup
}
\usepackage{fancyhdr}
\usepackage{caption}
\title{Technical Report: Competition Solution For BetterMixture}
\author{ Shuaijiang Zhao,Xiaoquan Fang \\
Beike Inc., Beijing, China  \\
\texttt{\{zhaoshuaijiang001,fangxiaoquan001\}@ke.com} }

\begin{document}
\maketitle
\begin{abstract}
In the era of flourishing large-scale models, the challenge of selecting and optimizing datasets from the vast and complex sea of data, to enhance the performance of large language models within the constraints of limited computational resources, has become paramount.
This paper details our solution for the BetterMixture challenge, which focuses on the fine-tuning data mixing for large language models. Our approach, which secured third place, incorporates data deduplication, low-level and high-level quality filtering, and diversity selection. 
The foundation of our solution is Ke-Data-Juicer \textsuperscript{1}\blfootnote{
\textsuperscript{1} https://github.com/shuaijiang/ke-data-juicer}, an extension of Data-Juicer, demonstrating its robust capabilities in handling and optimizing data for large language models.

\end{abstract}

\section{Introduction}
The emergence of large-scale language models such as ChatGPT\cite{chatgpt} has transformed  natural language processing. Meanwhile, the rapid growth of Chinese open-source large language models, including ChatGLM\cite{glm130b}, Baichuan\cite{yang2023baichuan}, Qwen\cite{qwen}, and BELLE\cite{BELLE},  contributing positively to the field's evolution.

The  swift development of Large Language Models (LLMs) has highlighted the critical need for vast quantities of high-quality data. 
In the response, BetterMixture emerges as a data-centric challenge that tests the analysis and combination capabilities of fine-tuning data for LLMs, bridging the gap between data needs and model optimization.

To tackle  the challenge, we utilized our Ke-Data-Juicer system, an advancement of Data-Juicer.
Data-Juicer\cite{chen2024datajuicer} is a comprehensive one-stop data processing system for Large Language Models. It is capable of efficiently generating a variety of data recipes, exploring numerous combinations for creating data mixtures, and assessing their impact on model performance. 
Ke-Data-Juicer builds upon this foundation by enhancing high-level quality filtering and diversity selection capabilities.

Building on Ke-Data-Juicer, we applied standard filtering techniques, including text length, language identification, and specific word filtering, referred to as low-level quality filtering.

To enhance data quality filtering, we introduced high-level quality filtering, with a LLM serving as a trainable data selector. This process evaluates and assigns scores to each sample of instruction fine-tuning data.
Specifically, we introduced Perplexity (PPL) calculated by the LLM to quantify the difficulty of instructions.
Instruction Following Difficulty (IFD) \cite{li2023quantity} also introduced to assess the challenge of responding to specific instructions. 
Furthermore, we introduced the IFD-Vote method, which utilizes multiple LLMs to refine quality assessment based on their collective scores.

In addition to quality, diversity is crucial. We employ the k-center-greedy algorithm to enhance the diversity of the selected data mixture

In summary, there are three main contributions of this paper:
\begin{itemize}
\item 
We proposed a complete solution for the BetterMixture challenge, securing third place in the competition.
\item
We introduced high-level quality filtering methods based on LLMs, including LLM perplexity filtering and LLM Instruction-Following Difficulty (IFD) filtering techniques.
\item
We introduced the IFD-Vote method, leveraging multiple LLMs, to select high-quality instruction data.
\end{itemize}

\section{BetterMixture Challenge}
BetterMixture\textsuperscript{1}\blfootnote{
\textsuperscript{1} https://tianchi.aliyun.com/competition/entrance/532174} is a data-centric challenge that assesses the ability to analyze and combine fine-tuning data for Large Language Models (LLMs). 
The organizers provide several candidate fine-tuning datasets, requiring participants to conduct data analysis, design mixing and sampling strategies, assign certain mixing ratios to each candidate subset, and create a mixed fine-tuning dataset within given computational constraints. 

The candidate data originate from 20 datasets of Alpaca-CoT. During data analysis and sampling, participants can only use these specified datasets and are strictly prohibited from modifying any data or adding external data.

This competition exclusively uses the Baichuan2-7B-Base model, employing PEFT (LoRA\cite{hu2021lora}) training with a training data limit of 10M tokens.

Evaluation is of paramount prominence to the accuracy and fairness of the competition. 
The evaluation dataset is detailed in Table\ref{details_eval}, encompasses a broad range of capabilities. 
The evaluation metrics involve computing the ratio of the participant's evaluation score on each task to the baseline score of the Baichuan2-7B-Base model. The leaderboard score is derived by averaging all the individual task ratios.

\section{Methodology}
Formally, we define the instruction dataset $X$ and employ a selection method $F$ to extract a subset $S$ from $X$, denoted as:
\begin{equation}
S = F(X)
\end{equation}

An evaluation metric Q to assess the quality of subset $S$, guiding to obtain the optimal selection method $F^\star$:
\begin{equation}
F^\star = \argmax_F Q(S)
\end{equation}

\begin{figure*}[t]
\centering
\setlength{\abovecaptionskip}{0.1cm} 
\includegraphics[width=1.0\textwidth, trim={1cm 7cm 1cm 6cm} ]{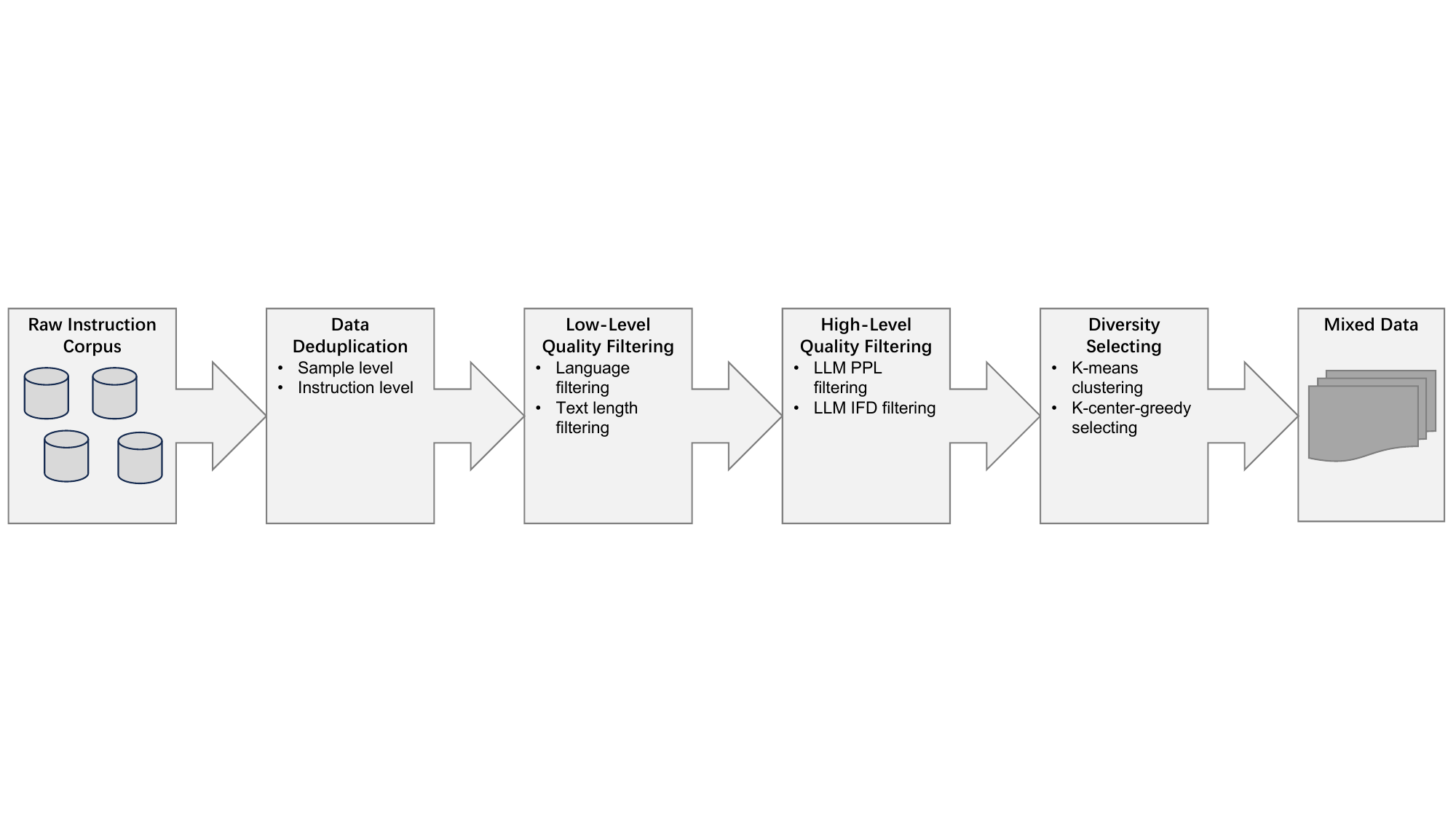}
\caption{Overview of our competition solution.}
\label{data_process}
\end{figure*}

This selection illustrates our optimal selection method $F^\star$, encompasses several critical steps: deduplication, quality filtering, and diversity selection.
The overview of our solution is shown in Figure\ref{data_process}.

\begin{figure*}[t]
\centering
\begin{multicols}{2}
    \includegraphics[width=\linewidth, trim={2cm 4cm 2cm 0cm}]{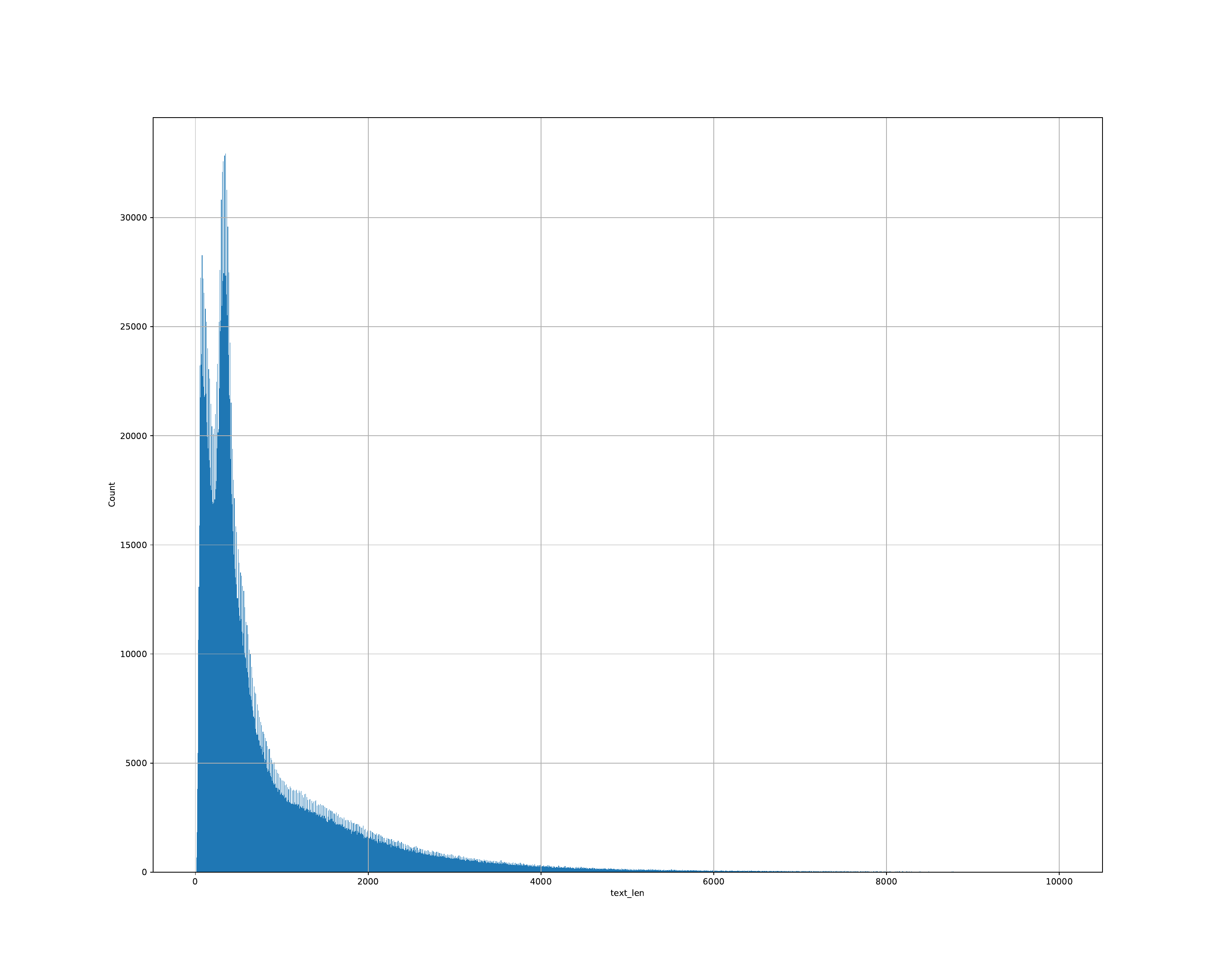}
    \caption*{(A) Text Length Distribution}
    
    \includegraphics[width=\linewidth, trim={2cm 4cm 2cm 0cm}]{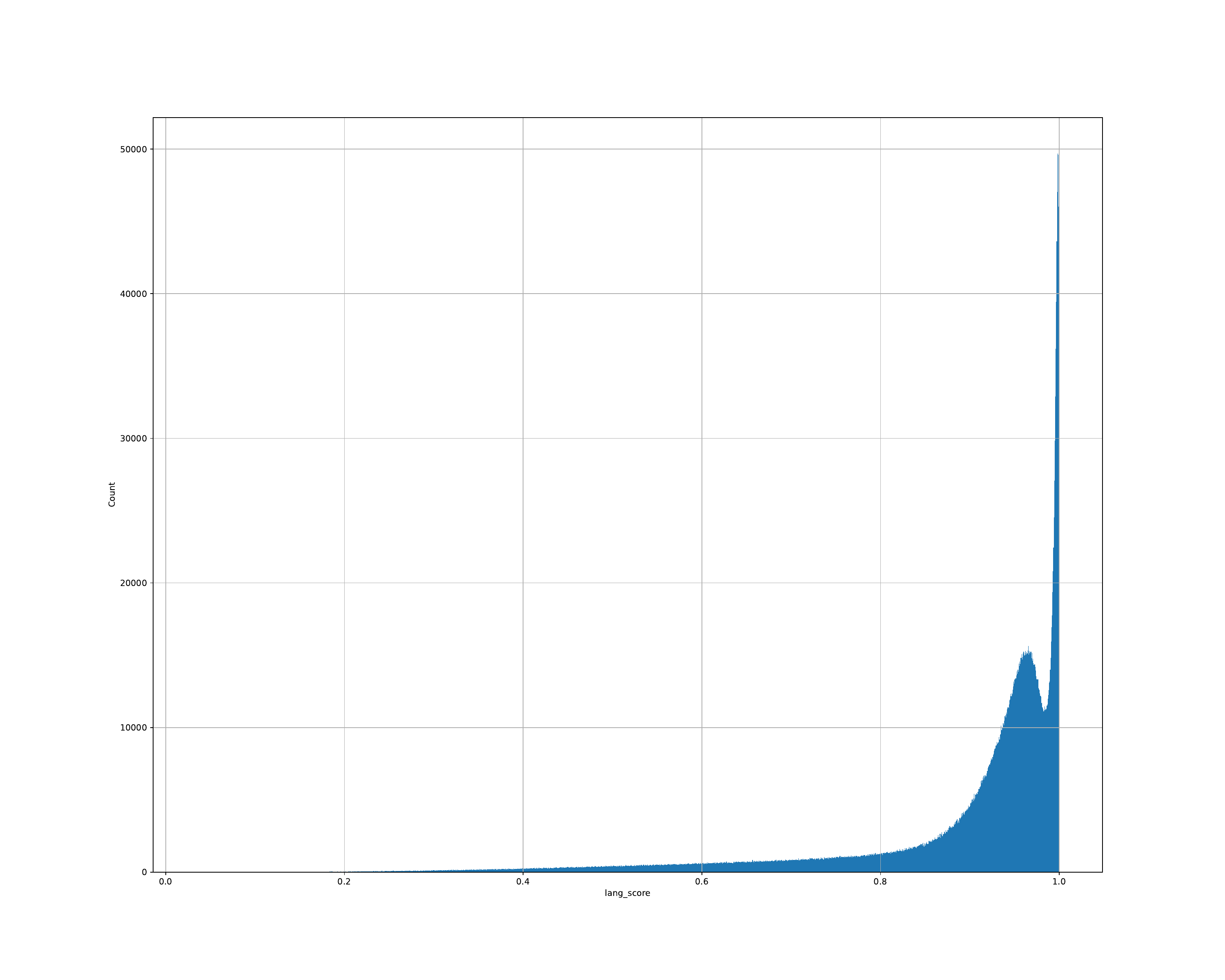}
    \caption*{(B) Language Identification Score Distribution}
    \end{multicols}
\begin{multicols}{2}
    \includegraphics[width=\linewidth,trim={2cm 4cm 2cm 1cm}]{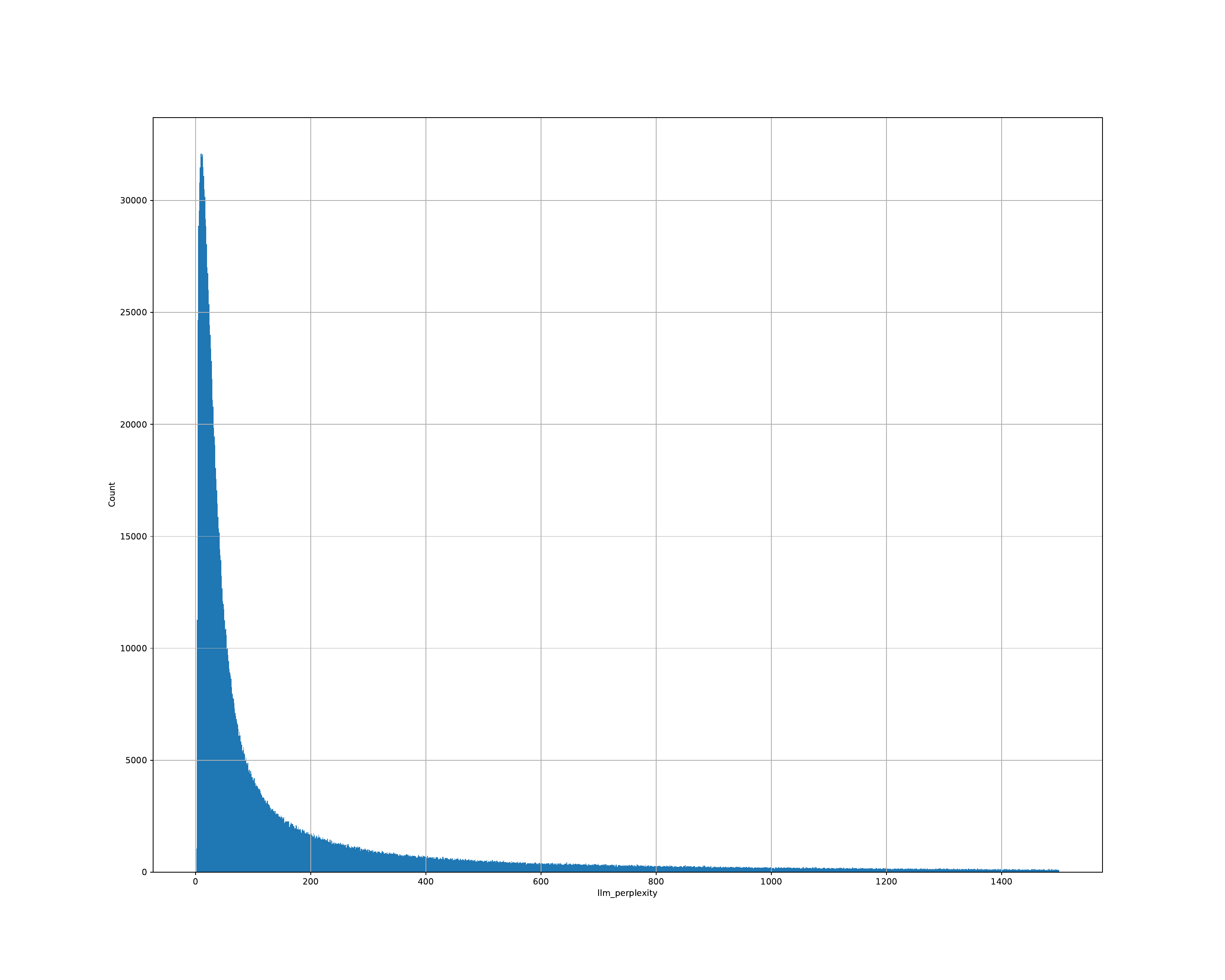}
    \caption*{(C) LLM Perplexity Distribution}

    \includegraphics[width=\linewidth,trim={2cm 4cm 2cm 1cm}]{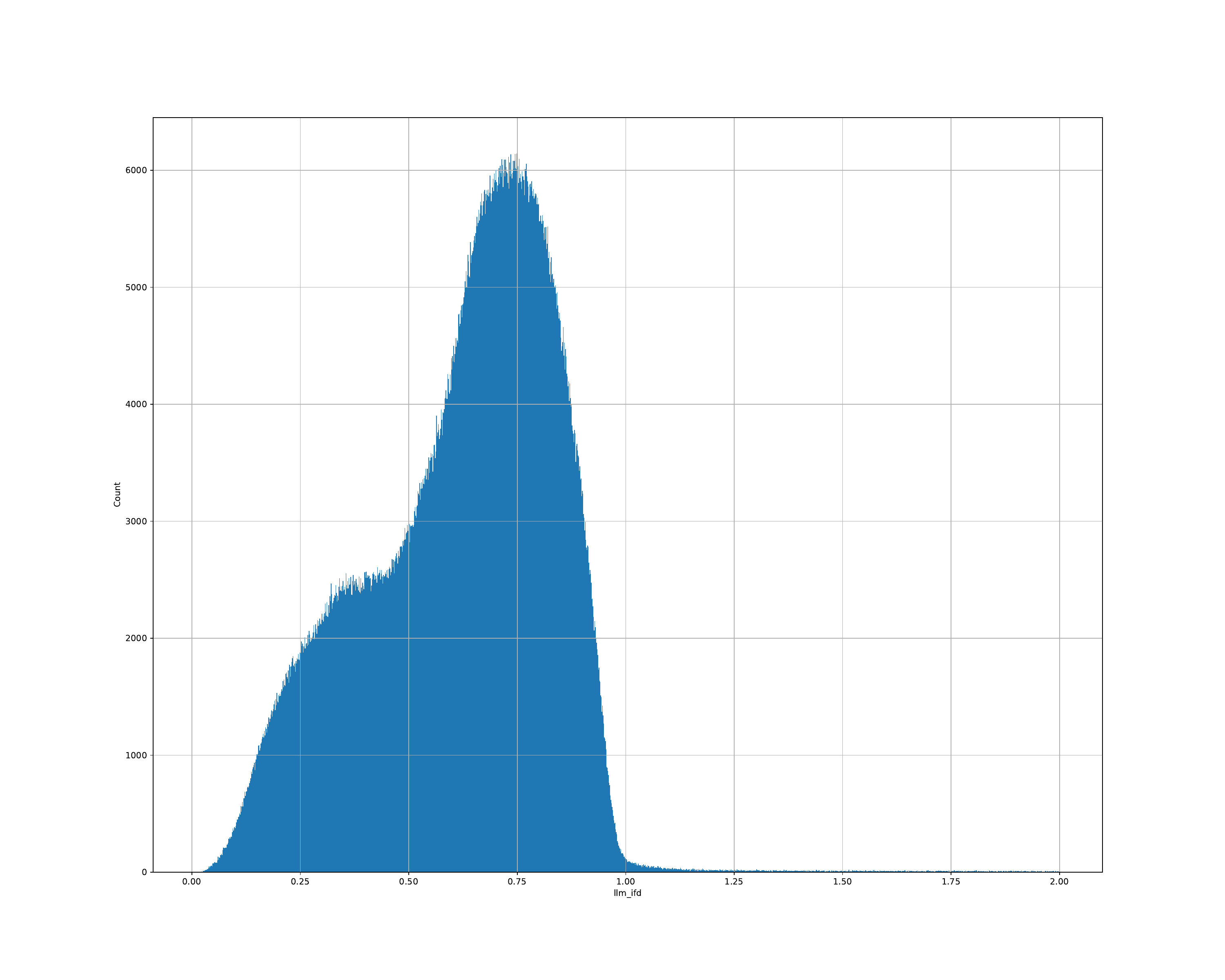}
    \caption*{(D) LLM IFD Score Distribution}
    \end{multicols}
\caption{Data statistics via histograms. (A) depicts the distribution of text lengths within the dataset. (B) shows the distribution of language identification scores. (C) presents the distribution of LLM perplexity. (D) illustrates the distribution of IFD scores.}
\label{origin_data}
\end{figure*}

\subsection{Data Deduplication}
\label{Data_Deduplication}
Data deduplication has become a fundamental process that boosts training efficiency and has the potential to improve the model's performance.
To avoid altering the data distribution, instead of hash based or model based deduplication approaches,
we opted for MD5 deduplication method via exact match.
This technique reduced the number of samples from 3.4 million to 2.7 million, streamlining dataset while preserving its diversity and richness.

\subsection{Low-level Quality Filtering}
After deduplication, we applied common standard filtering, namely low-level quality filtering, including text length and language identification filtering.
Through analyzing text lengths and language scores shown in Figure \ref{origin_data}, we established appropriate boundaries.

\textbf{Text Length Filtering} \quad We conducted statistical analysis on the length of samples, which are composed of instruction, input, and output, following data deduplication. 
We retained samples with text length ranging from 20 to 2000.

\textbf{Language Identification Filtering} \quad The dataset predominantly comprises English and Chinese languages, and the evaluation concentrates on English and Chinese. We preserved samples with English and Chinese with scores greater than 0.2.
Recognizing that certain types of samples, such as code and mathematics, might not achieve high language scores, we set the threshold score at 0.2 to accommodate these variations.

\subsection{High-level Quality Filtering}
To enhance data quality filtering, we employed LLM for scoring and filtering of the data, which refered as high-level quality filtering.
This approach encompasse LLM PPL, LLM IFD and LLM IFD-Vote filtering.
The data distribution after high-lebel quality filtering shown in Figure\ref{data_distribution} (B).

\textbf{LLM Perplexity filtering} \quad
Perplexity is a key evaluation metric for language models, measuring the model's understanding of instructions. 
The larger the perplexity value, the lower the model's understanding of the instructions, indicating that the model needs more of this data for training. 
However, to prevent the model from being affected by errors in the data itself, we control the perplexity within a targeted range of 20 to 1000. 
This avoids selecting data with excessively high perplexity, thereby preventing the introduction of excessive noise into the model.
We utilized the Baichuan2-7B-Base model, ensuring a consistent and reliable metric for filtering.

\textbf{LLM IFD filtering} \quad
IFD(Instruction Follow Difficulty) quantify the challenge each sample presents to the model, which is determined by the score of the conditional answer $S_\theta(A|Q)$ and the score of the model's direct answer $S_\theta(A)$. 
\begin{equation}
IFD = \frac{S_\theta(A|Q)}{S_\theta(A)}
\end{equation}
The score of the conditional answer measures the degree of consistency between the model's output and the correct answer corresponding to the instruction.
\begin{equation}
\begin{split}
&S_\theta(A|Q) =\\
&\frac{1}{N}\sum_{i=1}^{N}\log_{}{P(w_{i}^A|Q,w_{1}^A,w_{2}^A,...,w_{i-1}^A;\theta)}
\end{split}
\end{equation}
The direct answer score measures the inherent difficulty brought by the answer.
\begin{equation}
\begin{split}
S_\theta(A) =\frac{1}{N}\sum_{i=1}^{N}\log_{}{P(w_{i}^A|w_{1}^A,w_{2}^A,...,w_{i-1}^A;\theta)}
\end{split}
\end{equation}
The larger IFD score, the more difficult the instruction is, suggesting the model has more to learn from this data.
An IFD score too large or exceeding 1 suggests the instruction negatively impacts learning, whereas a low IFD score indicates the instruction is simple enough that the model can follow it without additional training. 

Therefore, to eliminate overly simplistic or anomalous data, we maintain the IFD score within the range of 0.2 to 0.9.
Adhering to the challenge's specifications, we employed the Baichuan2-7B-Base model to obtain the IFD score.

\textbf{LLM IFD-Vote filtering} \quad
As mentioned above, the IFD score indicating the challenge of a sample, is derived using the Baichuan2-7B-Base model in this study. However, due to the base model's limitations, the accuracy of the IFD score may vary. To enhance precision, we employed a fine-tuned version of the Baichuan2-7B-Base model as the IFD scorer. This model was fine-tuned with a mix of data processed through the previously and subsequently mentioned filtering and selection strategies.

Specifically, we calculated two IFD scores: one from the base model, another from the fine-tuned model. Samples exhibiting a variation of more than 50\% between  two scores were excluded.

\subsection{Diversity Selection}
Diversity is equally critical as quality to ensure the generalization capabilities of LLMs.
Meanwhile, given the constraint of training tokens capped at 10 million, corresponding to approximately 60,000 samples.
We select samples based on the IFD score and the k-center greedy algorithm to satisfy token constraint and guarantee high diversity.

\textbf{IFD based selection}
Specifically, we selected samples from each data source based on their average IFD score, targeting a total data size of 70,000. However, for some data sources, this target number was unmet, leading to a final collection of 60,000 samples. The details of data distribution shown in Figure \ref{data_distribution} (C).

\textbf{Language selection}
As illustrated in Table\ref{7bbase_result}, Baichuan2-7B-Base exhibits competitive performance with other similarly sized LLMs, even comparable to GPT3.5 Turbo which is a larger and more powerful LLM on Chinese benchmarks.
This suggests Baichuan2-7B-Base has already achieved a high level of proficiency in Chinese, offering limited room for further improvement through continuous training. 

To optimize quality and diversity, we applied k-center greedy algorithm to refine our selection of Chinese samples, reducing their number from 13,000 to 9,000 without performance degrade.

\section{Experiments}

\subsection{Baseline Models}
The baseline model required by the organizer is Baichuan2-7B-Base\cite{yang2023baichuan}, which was developed and released by Baichuan Intelligent Technology.

Baichuan2-7B-Base is a pre-trained model, boasting a parameter size of 7 billion and a training corpus comprising 2.6 trillion tokens.

\subsection{Dataset}
\begin{figure*}[t]
\centering
\begin{minipage}{0.5\textwidth}
    \includegraphics[width=\linewidth]{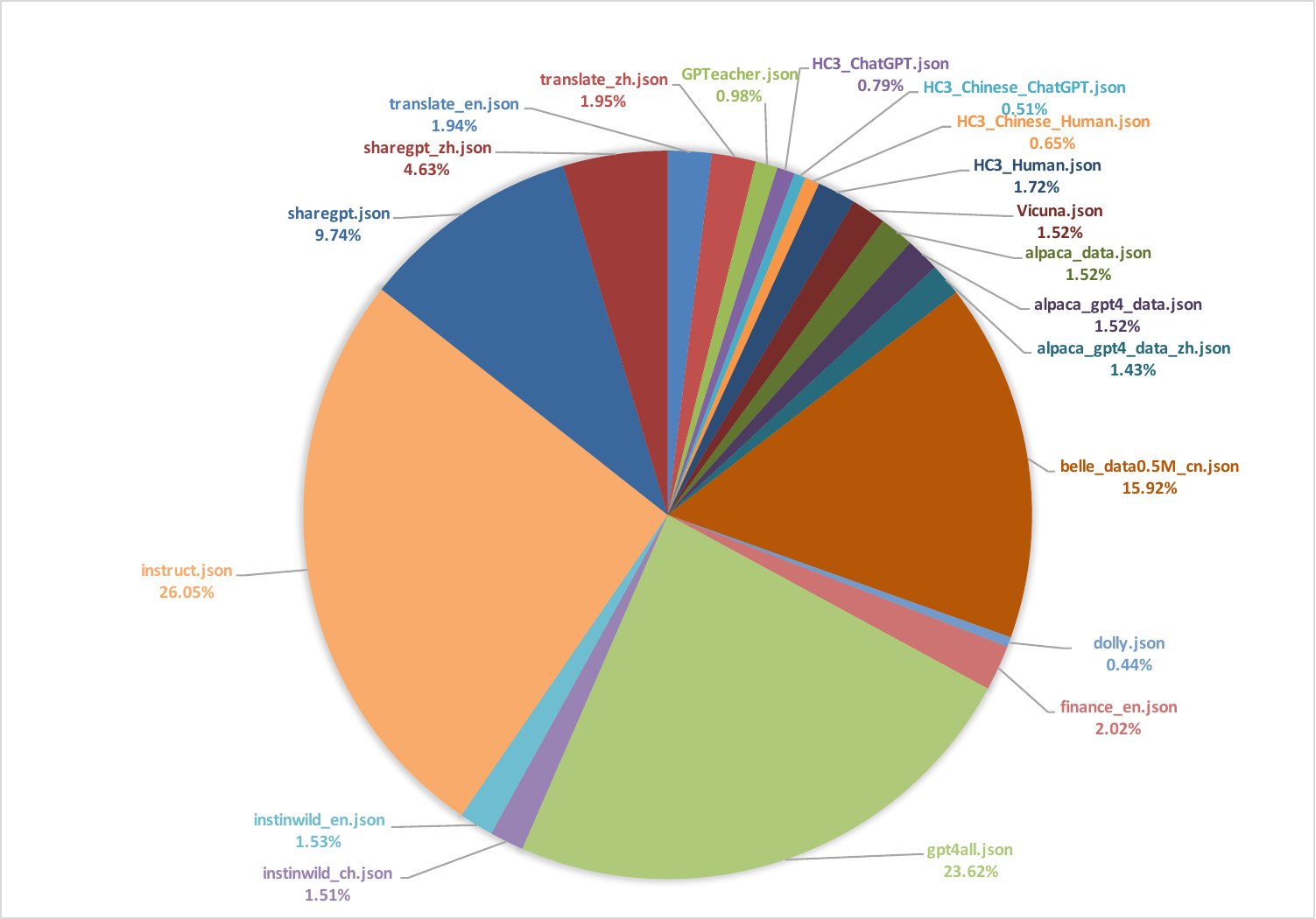}
    \caption*{(A) Distribution of original data}
\end{minipage}%
\begin{minipage}{0.5\textwidth}
    \includegraphics[width=\linewidth]{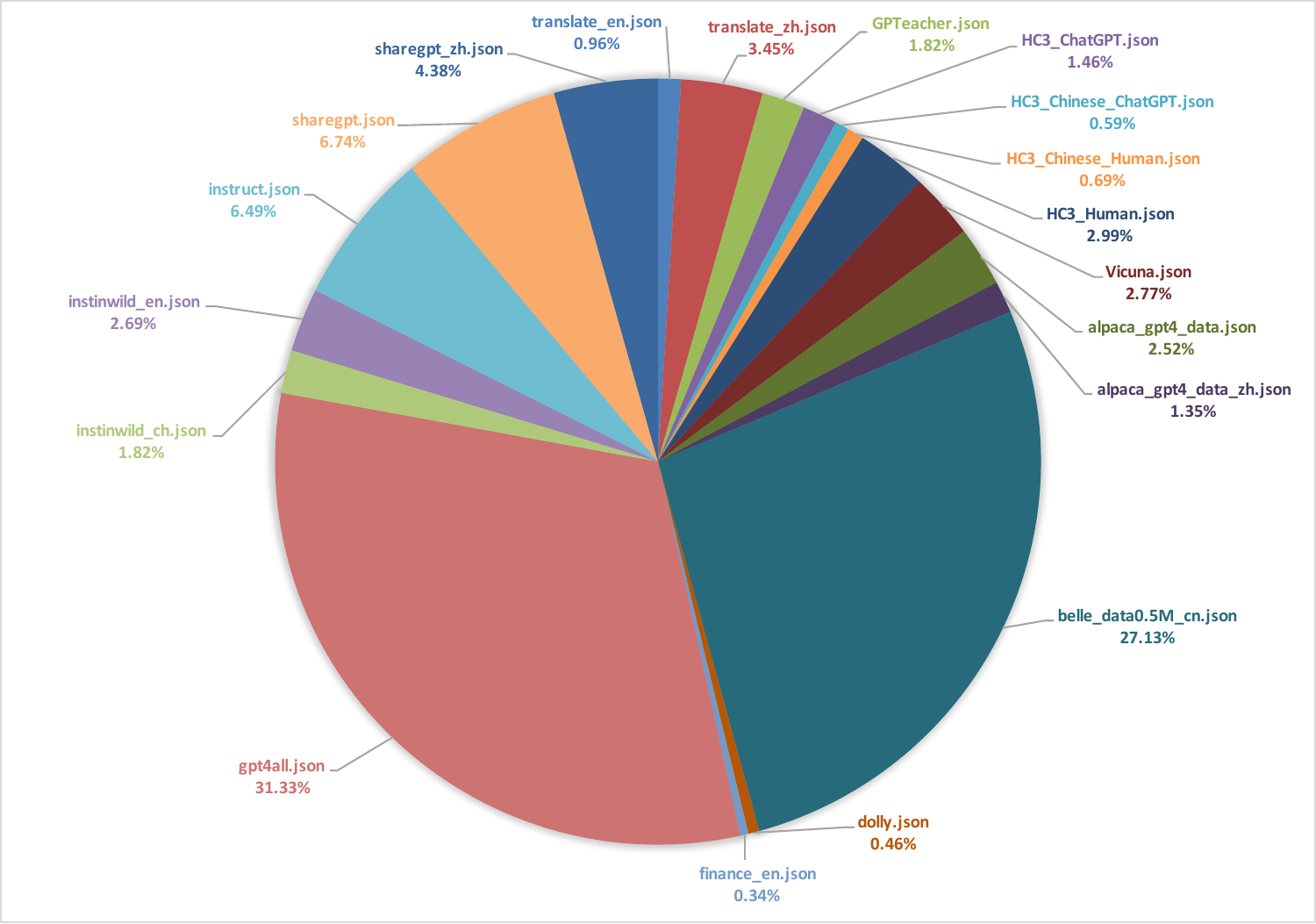}
    \caption*{(B) Distribution after applying filtering}
\end{minipage}
\begin{minipage}{0.5\textwidth}
    \includegraphics[width=\linewidth]{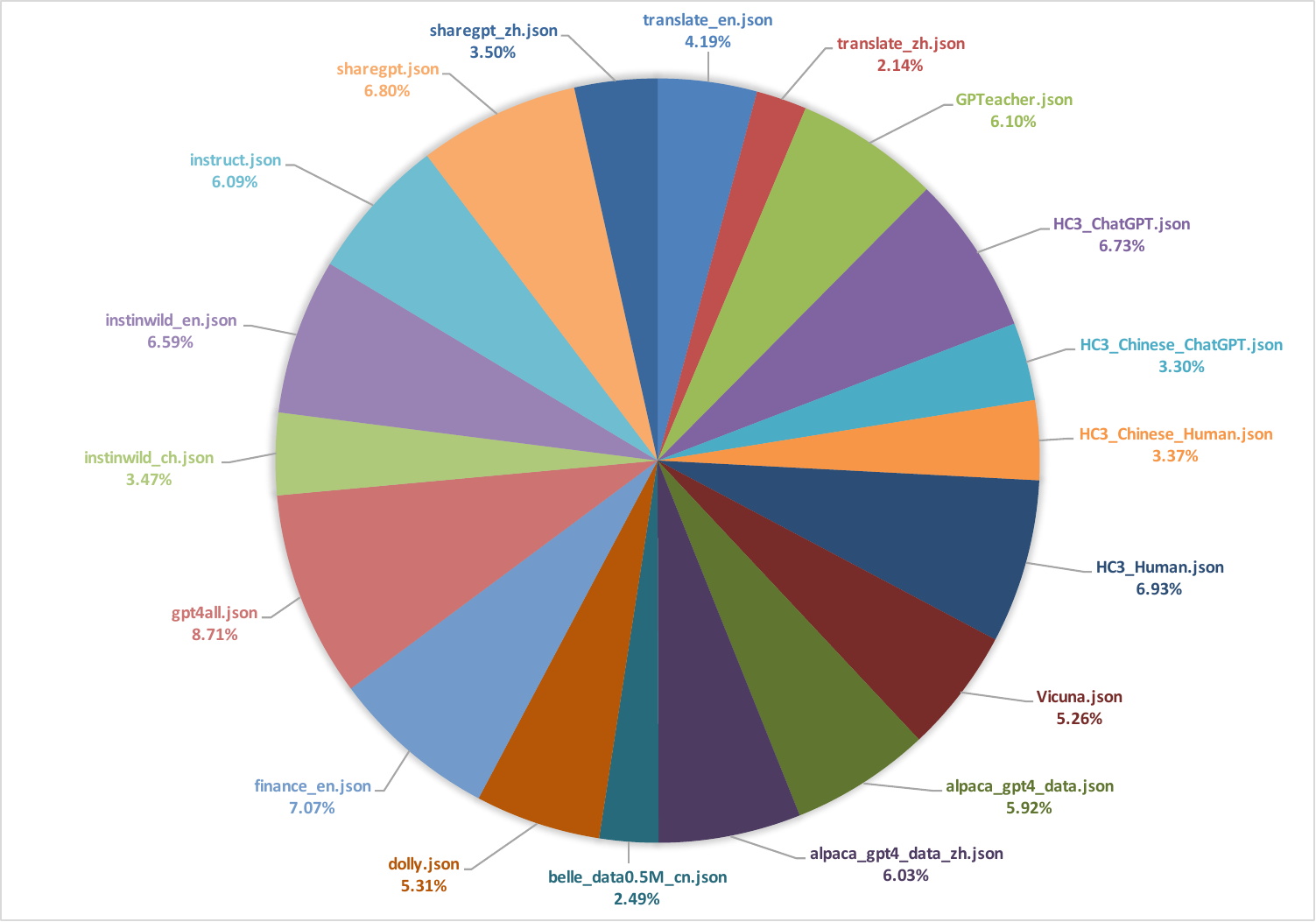}
    \caption*{(C) Distribution after IFD-based selection}
\end{minipage}%
\begin{minipage}{0.5\textwidth}
    \includegraphics[width=\linewidth]{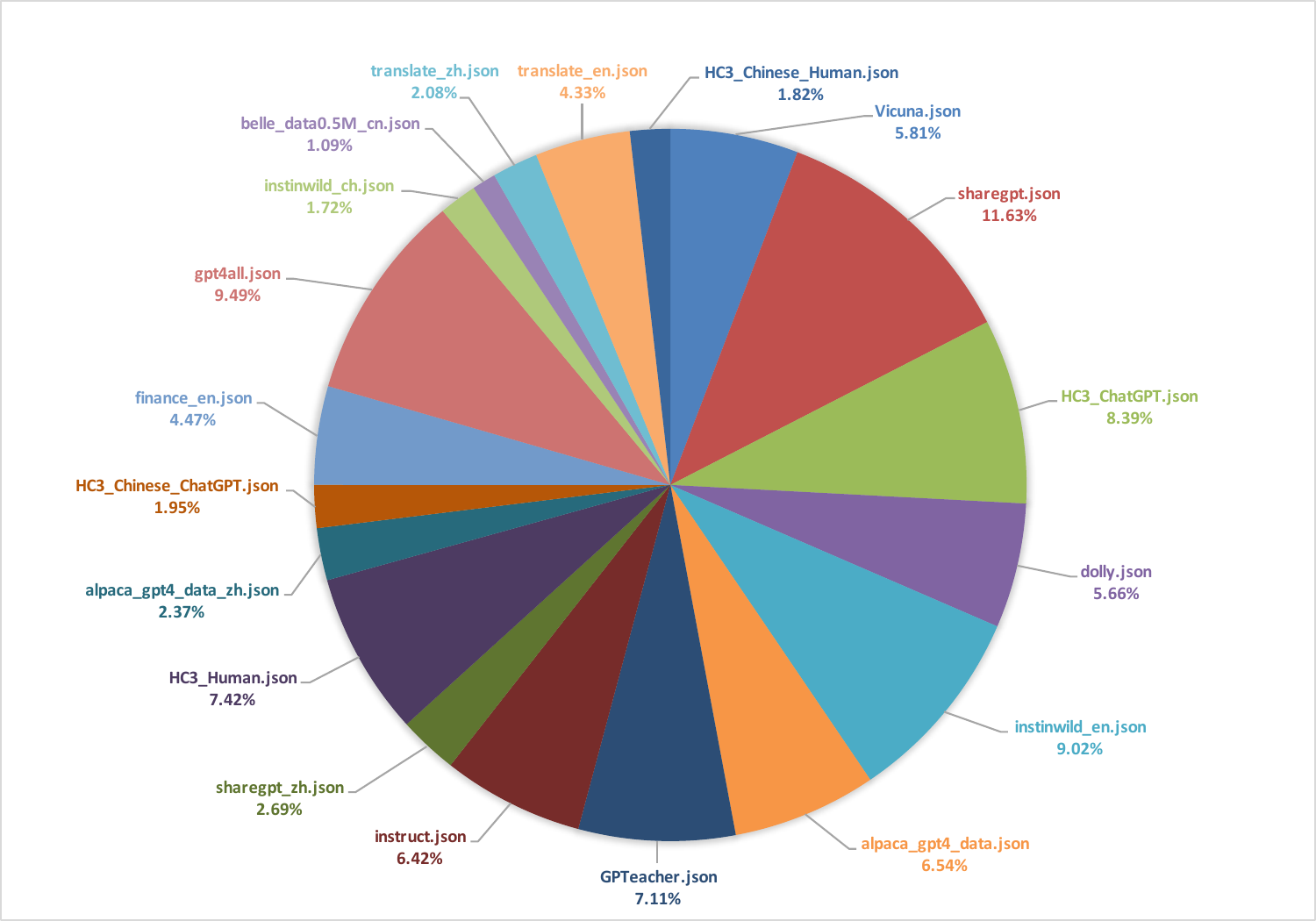}
    \caption*{(D) Distribution of final mixed data}
\end{minipage}
\caption{Data Distribution: (A) presents the original dataset distribution, (B) shows the distribution after applying low-level and high-level filtering, (C) depicts the distribution following IFD-based selection, and (D) illustrates the distribution of the final mixed data.}
\label{data_distribution}
\end{figure*}

We conducted a multi-stage analysis of the dataset: the original dataset, applying low-level and high-level filtering, following IFD-based selection, and concluding  the final dataset.
The data distribution for each of these stages is illustrated in Figure \ref{data_distribution}.

\subsection{Training Setups}
Most hyper-parameter settings were determined by the organizer.
The hyper-parameter settings we employed are detailed in Table \ref{hyper-parameters-ft}.
We chose a learning rate of 1e-3, the highest among 1e-3, 1e-4, and 1e-5.

\begin{table}[t]
\caption{Training Hyper-parameter settings}
\begin{center}
\begin{tabular}{l|c} 
\hline 
\textbf{Hyper parameter} & \textbf{Value} \\
\hline   
Precision  & bfloat16 \\
\hline
Epochs  & 3 \\
\hline
Batch size  & 1 \\
\hline
Learning rate  & 1e-3 \\
\hline
Warmup ratio  & 0.03 \\
\hline
LR scheduler type  & cosine \\
\hline
\end{tabular}
\end{center}
\label{hyper-parameters-ft}
\vspace{-20pt}
\end{table}
\subsection{Evaluation}
Table\ref{details_eval} presents a detailed overview of the evaluation stages for the competition, breaking down the specific categories assessed at each stage. 
In the preliminary stage, we outline the benchmarks for evaluating competitors' skills across a range of domains including reasoning, common sense, truthfulness, math, English knowledge, Chinese knowledge, and summarization. 
For the finals stage, while the domains remain the same, the datasets are undisclosed to participants.

\subsection{Results}
Table\ref{result} displays the evaluation scores, demonstrating the performance improvements achieved through various strategies.
Notably, the most effective improvements were attained through low-level and high-level filtering, IFD-selection, and k-center selection. 
Additionally, descending ordering samples by PPL and adopting a larger learning rate of 1e-3 proved to be beneficial.
\begin{table}[ht]
\centering
\caption{Evaluation Scores. Scores marked with a $\star$ represent the final stage, while others correspond to the preliminary stage.These two sets of scores are not directly comparable due to differing evaluation conditions.}
\begin{tabular}{c|p{4.5cm}c} 
\hline 
\textbf{Exp} & \textbf{ Strategies } & \textbf{Scores} \\
\hline 
\hline
Baseline  & random selection & 1.342  \\
\hline
Exp 1 &  low-level \& high-level filtering \& IFD-selection  &  1.401 \\
\hline
Exp 2& + k-center selection &  1.431 \\
\hline
Exp 3& + PPL descending order  &  1.443 \\
\hline
Exp 4& + Learning rate 1e-4 to 1e-3 &  1.455 \\
\hline
Exp 5& + IFD-Vote  &  $1.567^\star$ \\
\hline
\end{tabular}
\label{result}
\vspace{-20pt}
\end{table}

\begin{table*}[t]
\centering
\caption{Evaluation Detail}
\begin{center}
\begin{tabular}{c|cc} 
\hline 
\textbf{Category} & \textbf{ Preliminary Stage (\#Sample) } & \textbf{ Finals Stage} \\
\hline 
\hline
Reasoning & ARC(100)  & blind  \\
\hline
Common Sense & HellaSWAG(100)  & blind \\
\hline
Truthfulness  & TruthfulQA(100)  & blind \\
\hline
Math & GSM8K(100) & blind  \\
\hline
English Knowledge & MMLU(100*67) & blind\\
\hline
Chinese Knowledge & CMMLU(100*67) &blind \\
\hline
Summarization & SummScreen(100) & blind\\
\hline
\end{tabular}
\end{center}
\label{details_eval}
\end{table*}

\begin{table*}[t!]
\caption[Baichuan2-7B-Base] {Overall results\cite{yang2023baichuan} of Baichuan2-7B-Base compared with other similarly sized LLMs on general benchmarks}
\small
\begin{center}
\begin{tabular}{llcccccccc} 
\hline 
& \textbf{C-Eval} & \textbf{MMLU} & \textbf{CMMLU} & \textbf{Gaokao} & \textbf{AGIEval} & \textbf{BBH} & \textbf{GSM8K} & \textbf{HumanEval} \\
\hline
GPT-4 & 68.4 & 83.93 & 70.33 & 66.15 & 63.27 & 75.12 & 89.99 &69.51\\
GPT-3.5 Turbo & 51.10 & 68.54 & 54.06 & 47.07 & 46.13 & 61.59 & 57.77 & 52.44\\ 
\hline
\hline
\textbf{LlaMA-7B} & 27.10 & 35.10 & 26.75 & 27.81 & 28.17 & 32.38 & 9.78 & 11.59\\
 \textbf{LlaMA 2-7B} & 28.90 & 45.73 & 31.38 & 25.97 & 26.53 & 39.16 & 16.22 & 12.80 \\
\textbf{MPT-7B} & 27.15 & 27.93 & 26.00 & 26.54 & 24.83 & 35.20 & 8.64 & 14.02\\
\textbf{Falcon-7B} & 24.23 & 26.03 & 25.66 & 24.24 & 24.10 & 28.77 & 5.46 & - \\
\textbf{ChatGLM 2-6B} & 50.20 & 45.90 & 49.00 & \textbf{49.44} & \textbf{45.28} & 31.65 & \textbf{28.89} & 9.15\\
\textbf{Baichuan 1-7B} & 42.80 & 42.30 & 44.02 & 36.34 & 34.44 & 32.48 & 9.17 & 9.20\\
\textbf{Baichuan 2-7B-Base} & \textbf{54.00} & \textbf{54.16} & \textbf{57.07} & 47.47 & 42.73 & \textbf{41.56} & 24.49 & \textbf{18.29}\\
\hline
\end{tabular}
\end{center}
\label{7bbase_result}
\end{table*}

\section{Conclusions}
In this report, we introduce a solution for Better Mixture challenge, securing third place in the competition. 
We detailed the filtering and selection strategies implemented, highlighting their contribution to our success. Looking ahead, we aim to explore model-based data mixture learning techniques,such as DOREMI\cite{xie2024doremi}, as a promising direction for future work.
\bibliographystyle{acl}
\bibliography{acl2015}

\end{document}